\pdfoutput=1

\documentclass[11pt]{article}

\usepackage[preprint]{acl}

\usepackage{times}
\usepackage{algorithm}
\usepackage{algorithmic}
\usepackage{arydshln}
\usepackage{adjustbox}
\usepackage{amsmath}
\usepackage{mathtools}
\usepackage{amssymb}
\usepackage{amsfonts}
\usepackage{bm}
\usepackage{bbm}
\usepackage{booktabs}
\usepackage{multirow}
\usepackage{graphicx}
\usepackage{tabularx}
\usepackage{xcolor}
\usepackage{url}
\usepackage{siunitx}
\usepackage{booktabs}
\usepackage{makecell}
\usepackage{etoolbox}
\usepackage{caption}
\usepackage{subcaption}
\usepackage{comment}
\usepackage{latexsym}
\usepackage{mdframed}
\usepackage{float}

\usepackage{colortbl}    
\usepackage[table]{xcolor}

\definecolor{best}{RGB}{255,142,  0}      
\definecolor{secbest}{RGB}{255,199,115}   
\definecolor{secworst}{RGB}{178,210,255}  
\definecolor{worst}{RGB}{107,171,255}     

\newcommand{\best}[1]{\cellcolor{best}#1}
\newcommand{\secbest}[1]{\cellcolor{secbest}#1}
\newcommand{\secworst}[1]{\cellcolor{secworst}#1}
\newcommand{\worst}[1]{\cellcolor{worst}#1}

\usepackage[T1]{fontenc}

\usepackage[utf8]{inputenc}

\usepackage{microtype}

\usepackage{inconsolata}

\usepackage{graphicx}

\title{Decoding Uncertainty: The Impact of Decoding Strategies for \\ Uncertainty Estimation in Large Language Models}

\author{Wataru Hashimoto, \
  Hidetaka Kamigaito, \
  Taro Watanabe \\
  Nara Institute of Science and Technology \\
  \texttt{\{hashimoto.wataru.hq3, kamigaito.h, taro\}@is.naist.jp}}

\begin{document}
\maketitle
\begin{abstract}
Decoding strategies manipulate the probability distribution underlying the output of a language model and can therefore affect both generation quality and its uncertainty.
In this study, we investigate the impact of decoding strategies on uncertainty estimation in Large Language Models (LLMs).
Our experiments show that Contrastive Search, which mitigates repetition, yields better uncertainty estimates on average across a range of preference-aligned LLMs.
In contrast, the benefits of these strategies sometimes diverge when the model is only post-trained with supervised fine-tuning, i.e. without explicit alignment.
\end{abstract}

\section{Introduction}
\label{sec:introduction}
Recent advances in natural language processing (NLP) have been driven almost entirely by the rapid progress of Large Language Models (LLMs). State-of-the-art models such as GPT-4~\cite{openai2024gpt4technicalreport}, Llama~\cite{touvron2023llamaopenefficientfoundation}, and DeepSeek~\cite{deepseekai2025deepseekr1incentivizingreasoningcapability} already match or surpass human performance on a diverse suite of downstream NLP tasks.

Despite these successes, LLMs sometimes output fabricated or misleading text (hallucinations), which hinders the deployment of LLMs in safety-critical domains such as medicine, finance, and law.
Uncertainty Estimation (UE) is a key technique for mitigating the problem~\cite{geifman2018biasreduced,galil2023what,xin-etal-2021-art,hashimoto-etal-2024-data}.
By quantifying predictive uncertainty, a system can reject dubious outputs and route them to either human experts or stronger models.

In addition, decoding strategies also constitute a promising approach to address the problem.
Decoding strategies manipulate next-token distributions of language models, thereby able to elicit higher-quality outputs from the language model.
Recent work demonstrates that the choice of decoding strategy can markedly impact the quality of LLM outputs, underscoring its pivotal role in unlocking the full potential of these models~\cite{shi-etal-2024-thorough}.

However, the comprehensive investigation into how decoding strategies affect UE performance in LLMs remains limited.
Although recent studies improve uncertainty by devising sampling strategies~\cite{aichberger2025improving,vashurin2025cocoageneralizedapproachuncertainty}, the UE performance combined with extensive decoding strategies has not been systematically evaluated across various tasks.
Since decoding algorithms influence both the probability distribution over candidate tokens and the final token selection, they can have a significant impact on UE performance.
Furthermore, mainstream LLMs usually apply \textit{preference-alignment} techniques including Reinforcement Learning from Human Feedback (RLHF)~\cite{ouyang2022training} or Direct Preference Optimization (DPO)~\cite{rafailov2023direct} after Supervised Fine-Tuning (SFT).
Although such techniques improve the alignment of outputs with human preferences, recent work suggests they can degrade reliability~\cite{kadavath2022languagemodelsmostlyknow,openai2024gpt4technicalreport,tian-etal-2023-just,xiao2025restoring}, potentially interacting with the choice of decoding strategy.
The investigation of these interactions is therefore essential for producing more trustworthy LLM outputs.
We address this gap through two research questions:

\begin{itemize}
  \item RQ1: Which decoding strategies deliver the best UE performance?
  \item RQ2: How do training stages such as SFT and the preference-alignment techniques modulate UE performance across decoding strategies?
\end{itemize}

Our experiments reveal the following findings:
First, Contrastive Search, which explicitly mitigates repetition, achieves better UE performance as a whole.
Second, the optimal decoding strategy for UE can change as a model progresses from SFT to preference alignment during its post-training phase.
In addition, our results show that the changes in UE performance depend on the interaction with the decoding strategy and preference-alignment techniques.
All code is available at \url{https://github.com/wataruhashimoto52/decoding_uncertainty}.

\section{Decoding Strategies}
\label{sec:decoding_strategies}
We focus exclusively on deterministic decoding strategies because deterministic outputs are important in safety-critical domains such as finance~\cite{2024stochasticparrots,you2024trustsafetyllms}.
The strategies examined in this study are: Greedy Search (Greedy), Beam Search (BS)~\cite{freitag-al-onaizan-2017-beam}, Diverse Beam Search (DBS)~\cite{Vijayakumar_Cogswell_Selvaraju_Sun_Lee_Crandall_Batra_2018}, Contrastive Search (CS)~\cite{su2022a,su2023contrastive}, Contrastive Decoding (CD)~\cite{li-etal-2023-contrastive}, Frustratingly Simple Decoding (FSD; based on an $n$-gram model)~\cite{yang-etal-2024-frustratingly}, FSD‑vec (based on a vectorized n-gram model), Decoding by Contrastive Layers (DoLa)~\cite{chuang2024dola}, and Self-Logits Evolution Decoding (SLED)~\cite{zhang2024sled}.
Technical details and the hyper‑parameter search space are provided in Appendix~\ref{sec:appendix_details_decoding_strategies}.

\begin{table*}[t!]
\centering
\scalebox{0.43}{

\begin{tabular}{c|l|c|cc|ccc|c|c||c|cc|ccc|c|c}
\hline
\multicolumn{1}{c|}{\textbf{Model}} &
\multicolumn{1}{c|}{\textbf{Method}} &
\multicolumn{8}{c||}{\textbf{MSP}} &
\multicolumn{8}{c}{\textbf{MTE}} \\ \cline{3-18}

\multicolumn{1}{c|}{} & \multicolumn{1}{c|}{} &
\multicolumn{1}{c|}{TriviaQA} &
\multicolumn{2}{c|}{XSum} &
\multicolumn{3}{c|}{WMT19} &
\multicolumn{1}{c|}{HumanEval} &
\multicolumn{1}{c||}{} &
\multicolumn{1}{c|}{TriviaQA} &
\multicolumn{2}{c|}{XSum} &
\multicolumn{3}{c|}{WMT19} &
\multicolumn{1}{c|}{HumanEval} &
\multicolumn{1}{c}{} \\ \cline{3-18}
\multicolumn{1}{c|}{} & \multicolumn{1}{c|}{} &

\multicolumn{1}{c|}{RougeL} &
\multicolumn{1}{c}{RougeL} & \multicolumn{1}{c|}{AlignScore} &
\multicolumn{1}{c}{BLEU} & \multicolumn{1}{c}{Comet} & \multicolumn{1}{c|}{AlignScore} &
\multicolumn{1}{c|}{Pass@1} & \multicolumn{1}{c||}{Mean PRR} &
\multicolumn{1}{c|}{RougeL} &
\multicolumn{1}{c}{RougeL} & \multicolumn{1}{c|}{AlignScore} &
\multicolumn{1}{c}{BLEU} & \multicolumn{1}{c}{Comet} & \multicolumn{1}{c|}{AlignScore} &
\multicolumn{1}{c|}{Pass@1} & \multicolumn{1}{c}{Mean PRR} \\ \hline

\multirow{9}{*}{\rotatebox[origin=c]{90}{Llama2-7B-Chat}}
& Greedy & $62.97_{0.59}$ & $14.42_{1.87}$ & $1.57_{1.78}$ & $38.74_{2.40}$ & $46.48_{1.92}$ & $19.02_{3.14}$ & $-11.03_{8.61}$ & 20.95 & $49.13_{0.91}$ & $8.07_{1.60}$ & $10.68_{1.75}$ & $31.24_{2.21}$ & \secbest{$25.03_{1.93}$} & $21.69_{3.33}$ & $-13.49_{8.56}$ & 17.89 \\
& BS & $63.62_{0.70}$ & $14.12_{1.79}$ & $-0.45_{1.66}$ & \secbest{$38.96_{2.02}$} & $52.88_{1.56}$ & \secbest{$20.16_{3.56}$} & $-24.51_{7.04}$ & 18.65 & \secbest{$50.58_{0.85}$} & $4.68_{1.75}$ & $9.50_{1.69}$ & $29.78_{2.10}$ & $21.46_{2.05}$ & $18.11_{3.71}$ & \worst{$-28.47_{6.56}$} & 14.03 \\
& DBS & \best{$63.98_{2.34}$} & \secbest{$14.44_{1.71}$} & \worst{$-18.03_{1.42}$} & \secworst{$28.97_{2.41}$} & \secbest{$53.24_{1.82}$} & \secworst{$4.35_{3.66}$} & \secworst{$-35.36_{8.70}$} & \secbest{21.51} & $41.34_{0.94}$ & $7.50_{1.64}$ & \best{$12.54_{1.76}$} & \worst{$6.66_{2.34}$} & \worst{$-6.66_{2.27}$} & \secworst{$6.07_{3.44}$} & $-11.39_{8.38}$ & 14.25 \\
& CS & \secbest{$63.73_{0.69}$} & \best{$17.56^{*}_{1.04}$} & $1.43_{2.04}$ & $36.94_{2.29}$ & $41.99_{1.97}$ & $19.58_{3.21}$ & \secbest{$-9.96_{8.86}$} & \best{21.55} & \best{$51.95^{*}_{0.82}$} & \secbest{$8.99_{1.85}$} & $10.21_{2.14}$ & $30.66_{2.11}$ & $23.70_{2.09}$ & \secbest{$22.24_{3.13}$} & $-12.10_{8.28}$ & \secbest{18.66} \\
& CD & \secworst{$15.97_{2.22}$} & $7.18_{1.57}$ & \secworst{$-1.22_{1.19}$} & \best{$49.31^{*}_{1.43}$} & \best{$54.19^{*}_{1.34}$} & \best{$50.39^{*}_{2.14}$} & \best{$-7.20_{8.44}$} & 21.47 & \secworst{$24.95_{1.67}$} & \best{$9.33_{1.68}$} & \worst{$1.47_{2.04}$} & \best{$58.25^{*}_{1.27}$} & \best{$62.48^{*}_{1.26}$} & \best{$63.00^{*}_{1.87}$} & \best{$-5.67_{8.08}$} & \best{27.11} \\
& FSD & $33.84_{1.87}$ & \secworst{$0.64_{1.39}$} & \best{$12.42^{*}_{2.27}$} & $31.82_{1.74}$ & $14.04_{2.19}$ & $8.15_{3.58}$ & $-27.03_{8.42}$ & \secworst{9.97} & $34.81_{0.95}$ & \secworst{$-0.13_{1.38}$} & \secbest{$11.73_{2.26}$} & \secbest{$34.77_{1.77}$} & $15.93_{2.17}$ & $9.39_{3.55}$ & \secworst{$-27.03_{8.29}$} & \secworst{10.59} \\
& FSD-vec & $32.66_{1.81}$ & \worst{$-1.58_{1.36}$} & \secbest{$10.11_{2.14}$} & $31.24_{1.71}$ & \secworst{$12.83_{1.71}$} & $8.24_{3.45}$ & $-20.19_{8.71}$ & 16.81 & $32.08_{0.72}$ & \worst{$-3.04_{1.36}$} & $11.31_{2.01}$ & $33.54_{1.78}$ & \secworst{$14.21_{2.15}$} & $9.10_{3.44}$ & \secbest{$-9.14_{8.51}$} & 15.36 \\
& DoLa & $61.15_{0.65}$ & $12.46_{1.92}$ & $0.07_{1.56}$ & $38.78_{2.04}$ & $49.15_{1.79}$ & $16.74_{3.62}$ & $-14.82_{9.70}$ & 19.06 & $49.23_{0.85}$ & $5.51_{1.96}$ & \secworst{$8.17_{1.72}$} & $25.03_{1.89}$ & $17.36_{1.91}$ & $15.14_{3.67}$ & $-17.41_{9.02}$ & 14.28 \\
& SLED & \worst{$-12.68_{0.44}$} & - & - & \worst{$-27.36_{1.20}$} & \worst{$-61.37_{1.54}$} & \worst{$2.53_{3.04}$} & \worst{$-41.24_{8.58}$} & \worst{-19.69} & \worst{$7.80_{0.38}$} & - & - & \secworst{$16.24_{1.96}$} & $18.32_{2.06}$ & \worst{$5.34_{3.24}$} & $-17.63_{9.78}$ & \worst{2.94} \\
\hline
\multirow{6}{*}{\rotatebox[origin=c]{90}{Llama3-8B-RLHF}}
& Greedy & \secbest{$20.80_{0.86}$} & $15.23_{1.89}$ & $-2.44_{1.77}$ & \secbest{$59.6_{1.54}$} & \secbest{$82.27_{0.72}$} & \secbest{$20.49_{2.89}$} & $-30.76_{8.18}$ & \secbest{23.60} & \secbest{$20.27_{0.87}$} & \secbest{$9.78_{1.62}$} & $5.79_{1.93}$ & \secbest{$59.65_{1.37}$} & \secbest{$68.40_{1.01}$} & \secbest{$23.96_{2.75}$} & \secbest{$-31.33_{8.26}$} & \secbest{22.36} \\
& BS & \best{$22.88^{*}_{0.80}$} & \best{$38.32^{*}_{1.33}$} & \secbest{$2.64_{1.61}$} & $31.36_{1.46}$ & $48.11_{1.64}$ & \secworst{$-6.71_{4.23}$} & $-38.85_{5.46}$ & 13.96 & $19.31_{0.85}$ & \worst{$-1.54_{1.72}$} & \worst{$2.27_{1.85}$} & \secworst{$25.06_{2.20}$} & \worst{$8.10_{2.30}$} & \secworst{$16.87_{3.46}$} & \secworst{$-41.03_{4.87}$} & \secworst{4.15} \\
& DBS & \worst{$-9.70_{0.71}$} & \secworst{$12.03_{1.93}$} & \worst{$-9.52_{1.66}$} & \secworst{$29.8_{1.44}$} & \secworst{$24.84_{1.22}$} & $16.20_{3.92}$ & \best{$26.01^{*}_{10.17}$} & \secworst{12.81} & \worst{$11.68_{0.81}$} & $8.95_{1.82}$ & \secbest{$8.42_{2.09}$} & \worst{$13.48_{1.85}$} & \secworst{$15.77_{2.13}$} & \worst{$5.66_{4.65}$} & $-39.76_{5.40}$ & \worst{3.46} \\
& CS & \secworst{$16.89_{0.84}$} & \secbest{$16.37_{1.83}$} & $-2.51_{1.53}$ & \best{$61.24^{*}_{1.53}$} & \best{$84.86^{*}_{0.66}$} & $19.56_{3.10}$ & \secbest{$-28.34_{9.62}$} & \best{24.01} & $19.20_{0.79}$ & \best{$11.96^{*}_{1.57}$} & \secworst{$4.44_{1.73}$} & \best{$62.93^{*}_{1.38}$} & \best{$72.79^{*}_{0.98}$} & $22.54_{2.97}$ & \best{$-29.79_{8.63}$} & \best{23.44} \\
& CD & $17.81_{0.96}$ & \worst{$-9.18_{1.54}$} & \best{$9.90^{*}_{2.15}$} & \worst{$-9.03_{1.96}$} & \worst{$-57.07_{1.42}$} & \worst{$-34.46_{2.67}$} & \worst{$-51.43_{5.84}$} & \worst{-19.07} & \best{$22.25^{*}_{0.80}$} & \secworst{$-0.05_{1.84}$} & \best{$9.33^{*}_{1.83}$} & $41.62_{1.54}$ & $25.74_{1.67}$ & \best{$60.24^{*}_{2.04}$} & \worst{$-51.70_{5.85}$} & 15.35 \\
& DoLa & $20.11_{0.87}$ & $12.27_{2.02}$ & \secworst{$-3.29_{1.51}$} & $50.31_{1.82}$ & $76.96_{1.22}$ & \best{$21.13_{3.02}$} & \secworst{$-39.79_{5.85}$} & 19.67 & \secworst{$18.41_{0.80}$} & $6.29_{1.78}$ & $4.63_{1.77}$ & $52.17_{1.80}$ & $64.86_{1.11}$ & $19.90_{2.75}$ & $-40.46_{5.96}$ & 17.97 \\ \hline

\hline
\multirow{7}{*}{\rotatebox[origin=c]{90}{Zephyr-7B-$\beta$}}
& Greedy & \secbest{$64.85_{0.55}$} & $11.80_{2.05}$ & $-5.63_{1.79}$ & \secbest{$65.53_{1.53}$} & $81.67_{0.50}$ & $22.36_{3.31}$ & $-27.41_{5.56}$ & 30.45 & \secbest{$53.79_{0.77}$} & $10.80_{1.85}$ & $3.19_{2.09}$ & \secbest{$64.82_{1.46}$} & \secbest{$72.41_{0.89}$} & \best{$17.98^{*}_{3.63}$} & \secworst{$-35.24_{5.01}$} & \secbest{26.82} \\

& BS & $53.29_{1.18}$ & \secworst{$10.45_{1.92}$} & \secbest{$-5.32_{1.21}$} & $59.87_{1.71}$ & $80.85_{0.81}$ & $22.01_{3.41}$ & \worst{$-35.43_{4.71}$} & 26.53 & $44.48_{1.16}$ & \worst{$10.06_{1.73}$} & \worst{$2.13_{2.05}$} & $57.41_{1.57}$ & $62.98_{1.17}$ & \worst{$12.02_{3.96}$} & $-31.31_{3.67}$ & 22.54 \\

& DBS & $64.19_{0.61}$ & \worst{$1.50_{1.82}$} & \worst{$-17.84_{2.39}$} & $43.86_{1.45}$ & $75.24_{0.93}$ & $21.52_{3.16}$ & \best{$-21.22^{*}_{7.67}$} & 23.89 & $24.42_{0.88}$ & \secworst{$10.16_{1.57}$} & \best{$10.74^{*}_{1.94}$} & \worst{$36.38_{1.96}$} & $50.67_{1.49}$ & $16.28_{3.66}$ & \best{$-17.96^{*}_{6.92}$} & 18.67 \\

& CS & \best{$65.22^{*}_{0.51}$} & \best{$12.20^{*}_{1.97}$} & $-5.34_{1.83}$ & \best{$65.77_{1.47}$} & \best{$86.02^{*}_{0.58}$} & \best{$23.72^{*}_{3.30}$} & $-27.60_{7.18}$ & \best{31.43} & \best{$54.29^{*}_{0.77}$} & \secbest{$11.67_{1.84}$} & \secbest{$4.22_{2.16}$} & \best{$65.00^{*}_{1.45}$} & \best{$72.94^{*}_{1.06}$} & \secbest{$17.48_{3.49}$} & \secbest{$-28.63_{6.95}$} & \best{28.14} \\

& FSD & \secworst{$23.39_{0.83}$} & $11.27_{2.26}$ & \best{$2.04^{*}_{2.48}$} & \secworst{$42.28_{1.78}$} & \worst{$24.44_{1.76}$} & \worst{$11.52_{3.13}$} & \secworst{$-34.39_{5.28}$} & \worst{11.51} & \secworst{$23.87_{0.86}$} & $11.62_{2.23}$ & $3.32_{2.56}$ & $44.97_{1.76}$ & \worst{$27.66_{1.76}$} & \secworst{$12.54_{3.05}$} & \worst{$-36.03_{4.88}$} & \worst{12.56} \\

& FSD-vec & \worst{$23.21_{0.83}$} & \secbest{$11.82_{2.12}$} & \secworst{$-6.57_{2.54}$} & \worst{$41.23_{1.66}$} & \secworst{$26.66_{1.79}$} & \secworst{$15.39_{3.12}$} & $-28.12_{6.51}$ & \secworst{15.03} & \worst{$23.85_{0.86}$} & \best{$14.35^{*}_{2.15}$} & $2.79_{2.52}$ & \secworst{$43.84_{1.66}$} & \secworst{$28.64_{1.73}$} & $16.55_{2.98}$ & $-30.40_{6.49}$ & \secworst{14.23} \\

& DoLa & $63.90_{0.57}$ & $11.40_{2.04}$ & $-5.36_{1.78}$ & $65.22_{1.55}$ & \secbest{$84.54_{0.54}$} & \secbest{$22.81_{3.24}$} & \secbest{$-26.99_{5.69}$} & \secbest{30.79} & $52.02_{0.80}$ & $10.35_{1.84}$ & \secworst{$2.30_{2.08}$} & $60.00_{1.62}$ & $65.76_{1.12}$ & $15.82_{3.25}$ & $-28.72_{5.29}$ & 25.36 \\ \hline
\end{tabular}
}
\caption{PRRs for every task and generation metric pair in Llama2-7B-Chat, Llama3-8B-RLHF, and Zephyr-7B-$\beta$. Warmer color indicates better results. * indicates that the best strategy is significantly better (p < 0.05) than the second best. All standard deviations are obtained by bootstrap resampling with 1,000 trials.
}
\label{tab:rq1-prr-results}
\end{table*}

\section{Experimental Settings}
\label{sec:experimental_settings}
\subsection{Datasets}
We conducted evaluations across four text generation tasks: question answering (QA), text summarization (TS), machine translation (MT), and code generation (CG).
In QA, we use TriviaQA~\cite{joshi-etal-2017-triviaqa} dataset.
In TS, we use XSum~\cite{narayan-etal-2018-xsum} dataset.
In MT, we use WMT19~\cite{wmt19translate} dataset in German to English (De-En) setting.
In CG, we use HumanEval~\cite{chen2021evaluatinglargelanguagemodels} dataset. 
Dataset details are in Appendix~\ref{sec:appendix_dataset_details}.

\subsection{Models}
In RQ1, to examine the impact of decoding methods on UE performance across multiple tasks, we used Llama2-7B-Chat~\cite{touvron2023llama2openfoundation}\footnote{\url{https://huggingface.co/meta-llama/Llama-2-7b-chat-hf}}, Llama3-8B-RLHF~\cite{grattafiori2024llama3herdmodels, hu2024openrlhfeasytousescalablehighperformance}\footnote{\url{https://huggingface.co/OpenRLHF/Llama-3-8b-rlhf-100k}}, and Zephyr-7B-$\beta$~\cite{tunstall2024zephyr}.\footnote{\url{https://huggingface.co/HuggingFaceH4/zephyr-7b-beta}}
For CD, we adopted TinyLlama~\cite{zhang2024tinyllamaopensourcesmalllanguage}\footnote{\url{https://huggingface.co/TinyLlama/TinyLlama-1.1B-intermediate-step-955k-token-2T}} as the amateur model.
In RQ2, to evaluate the effects of the SFT and RLHF stages, we employed Llama3-8B-SFT\footnote{\url{https://huggingface.co/OpenRLHF/Llama-3-8b-sft-mixture}} and Llama3-8B-RLHF.
When applying preference tuning, we used Llama3-8B-DPO\footnote{\url{https://huggingface.co/RLHFlow/LLaMA3-iterative-DPO-final}}, which is applied the iterative version of DPO~\cite{dong2024rlhf}.
For CD, we adopted Llama3.2-1B-Instruct\footnote{\url{https://huggingface.co/meta-llama/Llama-3.2-1B-Instruct}} as the amateur.

\subsection{Details of Uncertainty Estimation}
\subsubsection{Uncertainty Estimation Metrics}
Following \citet{fadeeva-etal-2023-lm}, we measure UE performance with the Prediction–Rejection Ratio (PRR), which compares the area under the prediction–rejection curve obtained when ranking generations by model uncertainty to the oracle curve that ranks by true quality.
Unlike AUROC, PRR does not require binary labels, making it applicable to various text generation tasks such as TS or MT.
Let the test set be $\mathcal{D}=\{(\bm{x}_i,\bm{y}_i)\}$.
For each input $\bm{x}_i$, the language model produces an output $f(\bm{x}_i)$ and an associated uncertainty score $\mathcal{U}(\bm{x}_i)$.
The Prediction–Rejection Curve (PRC) traces the average min-max normalized generation quality $\mathcal{Q}(f(\bm{x}_i),\bm{y}_i)$ of those outputs that satisfy $\mathcal{U}(\bm{x}_i) < a$ as the rejection threshold $a$ varies.
The PRR compares the area under this curve when ranking by uncertainty against an oracle that ranks by true quality:

\begin{equation}
PRR \;=\; \frac{PRC_{\mathrm{uns}}}{PRC_{\mathrm{orc}}}.
\end{equation}

Here, $PRC_{\mathrm{orc}}$ is the area obtained when the lowest-quality samples are rejected first, whereas $PRC_{\mathrm{uns}}$ is the area when rejection is driven by the model’s uncertainty scores.
Because uncertainty is an imperfect proxy for quality, $PRC_{\mathrm{uns}}$ typically lies below $PRC_{\mathrm{orc}}$. A higher PRR means the uncertainty scores more accurately filter out low-quality outputs.\footnote{If correctly predicted instances receive higher uncertainty than mispredicted ones, the PRR can become negative.}

The quality score $\mathcal{Q}$ is task-dependent. 
The quality scores used in calculating the PRR are as follows: for QA we use RougeL \cite{lin-2004-rouge}; for TS we report RougeL and AlignScore~\cite{zha-etal-2023-alignscore}; for MT we report BLEU~\cite{papineni-etal-2002-bleu}, Comet~\cite{rei-etal-2020-comet} and AlignScore; for CG we report Pass@1~\cite{chen2021evaluatinglargelanguagemodels}.
To improve readability, all generation quality scores and PRRs are multiplied by 100.

\begin{table}[t!]
\scalebox{0.44}{

\begin{tabular}{c|l|c|cc|ccc|c}
\hline

\multicolumn{1}{c|}{\textbf{Model}} &
\multicolumn{1}{c|}{\textbf{Method}} &
\multicolumn{1}{c|}{TriviaQA} &
\multicolumn{2}{c|}{XSum} &
\multicolumn{3}{c|}{WMT19} &
\multicolumn{1}{c}{HumanEval} \\ \cline{3-9}
& & \multicolumn{1}{c|}{RougeL} &
\multicolumn{1}{c}{RougeL} & \multicolumn{1}{c|}{AlignScore} &
\multicolumn{1}{c}{BLEU} & \multicolumn{1}{c}{Comet} & \multicolumn{1}{c|}{AlignScore} &
\multicolumn{1}{c}{Pass@1} \\ \hline

\multirow{9}{*}{\rotatebox[origin=c]{90}{Llama2-7B-Chat}}
& Greedy      &   8.43   &   10.57   &   10.57   &   6.14    &   6.14    &   6.14    &   8.52 \\
& BS      &  24.95   &    7.41   &    7.13   &  16.26    &  16.26    &  20.30    &   6.26 \\
& DBS     & 1,751.65  & 2,020.93   & 2,012.05   & 1,971.69   & 1,971.69   & 1,977.82   & 2,604.76\\
& CS      &  17.42   &   10.61   &   10.61   &   8.16    &   8.16    &   6.11    &  12.77 \\
& CD      &   62.97     &  186.72   &  186.72   &  63.48    &  63.74    &  63.36    &  98.20 \\
& FSD     &   153.70     &   15.52   &   16.25   &   8.96    &   8.96    &   8.96    &   7.56 \\
& FSD-vec &   92.10     &   15.57   &   16.99   &   9.09    &   9.09    &   8.92    & 102.44 \\
& DoLa    &   6.32   &    7.54   &    7.33   &   4.82    &   4.82    &   4.82    &   6.69 \\
& SLED    &   -1,720.95     & -1,959.09  & -2,441.84  & -2,669.11   & -2,135.69   & -2,669.11   & -4,218.55 \\ \hline
\multirow{6}{*}{\rotatebox[origin=c]{90}{Llama3-8B-RLHF}}
& Greedy & 34.76 & 9.00 & 9.00 & 9.00 & 23.35 & 23.35 & 17.13\\
& BS     & 90.52 & 351.83 & 351.83 & 351.83 & 187.62 & 191.84 & 8.24\\
& DBS    & 2249.43 & 1210.59 & 1175.13 & 1213.68 & 1537.93 & 1609.42 & 2488.23\\
& CS     & 50.10 & 9.58 & 9.58 & 9.58 & 23.42 & 24.19 & 26.49 \\
& CD     & 47.81 & 175.62 & 175.62 & 175.62 & 374.31 & 374.31 & 10.68\\
& DoLa   & 24.43 & 4.91 & 4.89 & 4.89 & 15.33 & 15.33 & 10.63\\ \hline
\multirow{7}{*}{\rotatebox[origin=c]{90}{ Zephyr-7B-$\beta$}}
& Greedy   & 15.79   & 20.70   & 20.70   & 11.12   & 11.12   & 11.12   & 8.11  \\
& BS       & 41.49   & 17.81   & 16.91   & 19.47   & 19.47   & 19.47   & 5.17  \\
& DBS      & 1,470.62 & 1,958.16 & 1,958.16 & 1,276.36 & 1,276.36 & 1,395.72 & 2,210.45 \\
& CS       & 16.03   & 21.45   & 21.45   & 11.46   & 19.08   & 11.46   & 9.58  \\
& FSD      & 23.29   & 23.30   & 23.30   & 18.69   & 20.23   & 20.23   & 9.73  \\
& FSD-vec  & 23.28   & 23.62   & 23.62   & 18.62   & 20.25   & 20.25   & 11.72 \\
& DoLa     & 12.63   & 16.73   & 16.73   & 9.01    & 9.01    & 9.30    & 6.67  \\ \hline
\end{tabular}
}
\caption{Averaged MSP scores for every task and generation metric pair in Llama2-7B-Chat, Llama3-8B-RLHF, and Zephyr-7B-$\beta$. Higher score indicates more uncertain.}
\label{tab:rq1-msp-scores}
\end{table}

\begin{table}[t!]
\scalebox{0.45}{

\begin{tabular}{c|l|c|cc|ccc|c}
\hline

\multicolumn{1}{c|}{\textbf{Model}} &
\multicolumn{1}{c|}{\textbf{Method}} &
\multicolumn{1}{c|}{TriviaQA} &
\multicolumn{2}{c|}{XSum} &
\multicolumn{3}{c|}{WMT19} &
\multicolumn{1}{c}{HumanEval} \\ \cline{3-9}
& & \multicolumn{1}{c|}{RougeL} &
\multicolumn{1}{c}{RougeL} & \multicolumn{1}{c|}{AlignScore} &
\multicolumn{1}{c}{BLEU} & \multicolumn{1}{c}{Comet} & \multicolumn{1}{c|}{AlignScore} &
\multicolumn{1}{c}{Pass@1} \\ \hline
\multirow{8}{*}{\rotatebox[origin=c]{90}{Llama2-7B-Chat}}
& Greedy      &   0.13   &    0.16   &    0.16   &   0.25    &   0.25    &   0.25    &   0.10 \\
& BS      &   0.10   &    0.32   &    0.26   &   0.23    &   0.23    &   0.23    &   0.08 \\
& DBS     &   0.21   &    0.37   &    0.35   &   0.16    &   0.16    &   0.25    &   0.19 \\
& CS      &   0.13   &    0.40   &    0.40   &   0.25    &   0.25    &   0.25    &   0.12 \\
& CD      &   0.25   &    0.54   &    0.54   &   0.16    &   0.27    &    6.03   &   0.09 \\
& FSD     &   0.11   &    0.60   &    0.60   &   0.25    &   0.25    &   0.25    &   0.09 \\
& FSD-vec &   0.24   &    0.55   &    0.55   &   0.25    &   0.25    &   0.25    &   0.09 \\
& DoLa    &   0.05   &    0.19   &    0.23   &   0.18    &   0.18    &   0.18    &   0.05 \\ \hline

\multirow{6}{*}{\rotatebox[origin=c]{90}{Llama3-8B-RLHF}}
& Greedy & 0.80 & 0.36 & 0.36 & 0.36 & 0.89 & 0.89 & 0.39\\
& BS     & 0.52 & 0.27 & 0.27 & 0.27 & 0.54 & 0.51 & 0.18\\
& DBS    & 0.48 & 0.46 & 0.39 & 0.34 & 0.59 & 0.60 & 0.44\\
& CS     & 0.75 & 0.37 & 0.37 & 0.37 & 0.88 & 0.88 & 0.45\\
& CD     & 0.85 & 0.74 & 0.74 & 0.74 & 0.76 & 0.76 & 0.24\\
& DoLa   & 0.41 & 0.19 & 0.19 & 0.19 & 0.42 & 0.42 & 0.18\\ \hline
\multirow{7}{*}{\rotatebox[origin=c]{90}{Zephyr-7B-$\beta$}}
& Greedy   & 0.44 & 0.50 & 0.50 & 0.34 & 0.34 & 0.34 & 0.18 \\
& BS       & 0.30 & 0.41 & 0.38 & 0.25 & 0.25 & 0.25 & 0.11 \\
& DBS      & 0.38 & 0.45 & 0.45 & 0.29 & 0.29 & 0.33 & 0.27 \\
& CS       & 0.44 & 0.49 & 0.49 & 0.34 & 0.34 & 0.34 & 0.20 \\
& FSD      & 0.52 & 0.51 & 0.51 & 0.42 & 0.44 & 0.44 & 0.21 \\
& FSD-vec  & 0.52 & 0.51 & 0.51 & 0.42 & 0.43 & 0.43 & 0.22 \\
& DoLa     & 0.26 & 0.31 & 0.31 & 0.21 & 0.21 & 0.21 & 0.12 \\ \hline
\end{tabular}
}
\caption{Averaged MTE scores for every task and generation metric pair in Llama2-7B-Chat, Llama3-8B-RLHF, and Zephyr-7B-$\beta$. Higher score indicates more uncertain.}

\label{tab:rq1-mte-scores}
\end{table}

\subsubsection{How to Estimate Uncertainty Score}
To convert the predictive token‑level probability distribution into a single uncertainty score, an aggregation scheme must be chosen. 
To analyze the impact of decoding strategies on predictive uncertainty from the viewpoint of probability and entropy, we limit our analysis to two fundamental methods: Maximum Sequence Probability (MSP) which is the negative log‑likelihood of the generated sequence, and Mean Token Entropy (MTE) which is the average entropy of the token‑level predictive distributions.\footnote{More advanced methods can affect uncertainty estimates combined with the choice of the decoding strategy. Results obtained by applying one such technique -- Shifting Attention to Relevance~\cite{duan-etal-2024-shifting} -- to the computation of uncertainty scores are presented in Appendix~\ref{sec:appendix_advanced_ue_method}.}
For each decoding strategy tested, MSP and MTE are computed, and the resulting uncertainty scores are then evaluated using PRRs.

\section{Results \& Analysis}
\label{sec:results_analysis}
\subsection{RQ1: Which decoding strategies deliver the best UE performance?}
Table~\ref{tab:rq1-prr-results} reports the PRRs obtained with MSP and MTE when each of the decoding strategies is applied across four benchmarks.
In addition, Table~\ref{tab:rq1-msp-scores} and Table~\ref{tab:rq1-mte-scores} report averaged MSP scores and MTE scores, respectively.

\paragraph{Contrastive Search shows better uncertainty across the models on average.} 
Across all aligned models examined, CS, followed by Greedy, produces better uncertainty, on average.
We hypothesized that these results are due to CS's ability to mitigate repetition which is one of the causes of overconfidence in a language model~\cite{Holtzman2020The} while keeping the original probability~\cite{su2022a,su2023contrastive}.
To evaluate this, we measured averaged sentence-level Distinct-$n$~\cite{li-etal-2016-diversity}.
Distinct-$n$ is the rate at which $n$-grams in the output are different, which can evaluate the diversity of tokens in the output sentences.
As shown in Table~\ref{tab:distinct-n}, CS has the highest Distinct-1 and Distinct-2 overall, suggesting that the outputs from CS have less repetition than other decoding strategies.

\begin{table}[t!]
\scalebox{0.42}{

\begin{tabular}{c|l|rrrr|rrrr}
\hline
\multicolumn{1}{c|}{\textbf{Model}} &
\multicolumn{1}{c|}{\textbf{Method}} &
\multicolumn{4}{c|}{\textbf{Distinct-1}} &
\multicolumn{4}{c}{\textbf{Distinct-2}} \\ \cline{3-10}
\multicolumn{1}{c|}{} &
\multicolumn{1}{c|}{} &
\multicolumn{1}{c}{TriviaQA} &
\multicolumn{1}{c}{XSum} &
\multicolumn{1}{c}{WMT19} &
\multicolumn{1}{c|}{HumanEval} & 
\multicolumn{1}{c}{TriviaQA} &
\multicolumn{1}{c}{XSum} &
\multicolumn{1}{c}{WMT19} &
\multicolumn{1}{c}{HumanEval} \\ \hline

\multirow{9}{*}{\rotatebox[origin=c]{90}{Llama2-7B-Chat}}
& Greedy & 0.750 & 0.809 & 0.784 & 0.608 & 0.924 & 0.974 & 0.935 & 0.851 \\
& BS & 0.744 & 0.808 & 0.775 & 0.602 & 0.919 & 0.973 & 0.929 & 0.845  \\
& DBS & 0.739 & 0.806 & 0.766 & 0.598 & 0.917 & 0.973 & 0.925 & 0.842  \\
& CS & 0.759 & 0.819 & 0.807 & 0.604 & 0.930 & 0.977 & 0.948 & 0.846  \\
& CD & 0.739 & 0.782 & 0.729 & 0.569 & 0.909 & 0.950 & 0.887 & 0.801  \\
& FSD & 0.735 & 0.773 & 0.716 & 0.573 & 0.910 & 0.946 & 0.884 & 0.808  \\
& FSD-vec & 0.734 & 0.766 & 0.706 & 0.556 & 0.907 & 0.942 & 0.881 & 0.785  \\
& DoLa & 0.748 & 0.771 & 0.716 & 0.560 & 0.923 & 0.946 & 0.888 & 0.792  \\
& SLED & 0.744 & - & 0.716 & 0.554 & 0.919 & - & 0.889 & 0.790   \\
\hline
\multirow{6}{*}{\rotatebox[origin=c]{90}{Llama3-8B-RLHF}}
& Greedy & 0.703 & 0.805 & 0.766 & 0.656 & 0.914 & 0.941 & 0.861 & 0.885  \\
& BS & 0.692 & 0.804 & 0.766 & 0.648 & 0.909 & 0.941 & 0.861 & 0.881  \\
& DBS & 0.677 & 0.783 & 0.735 & 0.642 & 0.899 & 0.924 & 0.828 & 0.877  \\
& CS & 0.740 & 0.866 & 0.864 & 0.666 & 0.943 & 0.991 & 0.965 & 0.890  \\
& CD & 0.687 & 0.750 & 0.689 & 0.650 & 0.905 & 0.883 & 0.789 & 0.882  \\
& DoLa & 0.678 & 0.766 & 0.678 & 0.644 & 0.907 & 0.901 & 0.799 & 0.881  \\ \hline
\multirow{7}{*}{\rotatebox[origin=c]{90}{Zephyr-7B-$\beta$}}
& Greedy & 0.734 & 0.778 & 0.755 & 0.545 & 0.920 & 0.963 & 0.887 & 0.792 \\
& BS & 0.754 & 0.776 & 0.782 & 0.531 & 0.919 & 0.960 & 0.896 & 0.772 \\
& DBS & 0.747 & 0.775 & 0.771 & 0.538 & 0.919 & 0.960 & 0.890 & 0.787 \\
& CS & 0.734 & 0.782 & 0.774 & 0.557 & 0.922 & 0.966 & 0.906 & 0.808 \\
& FSD & 0.666 & 0.762 & 0.648 & 0.544 & 0.895 & 0.957 & 0.827 & 0.793 \\
& FSD-vec & 0.665 & 0.760 & 0.644 & 0.546 & 0.892 & 0.956 & 0.816 & 0.797 \\
& DoLa & 0.734 & 0.778 & 0.753 & 0.545 & 0.920 & 0.963 & 0.886 & 0.792 \\
\hline
\end{tabular}
}
\caption{Distinct-1 and Distinct-2 for every task and generation metric pair in Llama2-7B-Chat, Llama3-8B-RLHF, and Zephyr-7B-$\beta$. Higher score indicates diversified outputs.}
\label{tab:distinct-n}
\end{table}

\paragraph{BS and DBS sometimes underperform.}
We can see that BS and DBS in Llama3-8B-RLHF and Zephyr-7B-$\beta$ sometimes perform worse.
In Tables~\ref{tab:rq1-msp-scores} and Table~\ref{tab:rq1-mte-scores}, the negative log-probability and the entropy change significantly with BS and DBS compared to Greedy or CS.
The uncertainty scores obtained based on the manipulated probability distributions by BS and DBS may not be aligned with the objective of separating high- and low-quality outputs.

\paragraph{For CD, there is a large difference in UE performance across models.}
On average, CD provides the strongest aggregate performance, followed by CS in Llama2-7B-Chat.
However, the advantage of CD is mainly pronounced in MT setting.
On the other hand, in Llama3-8B-RLHF, the situation is markedly different: while CD remains reasonably reliable for factual metrics, its reliability deteriorates sharply for MT or TS.
As CD is highly sensitive to the specific pairing of teacher and student models, substantial behavioral differences across model families can be expected. Results of other models also support this in Appendix~\ref{sec:appendix_results_qwen2.5} and Appendix~\ref{sec:appendix_results_larger_model}.
Moreover, as with BS and DBS, we can see that the probability distribution of the LLM output changes significantly when CD is used, from Table~\ref{tab:rq1-msp-scores} and Table~\ref{tab:rq1-mte-scores}.
The selection and construction of an appropriate amateur model for CD to optimize UE performance remains an open challenge.

\paragraph{Recent factuality decoding strategies underperform.}
As shown in Table~\ref{tab:rq1-prr-results}, recent factuality decoding strategies such as DoLa and SLED frequently underperform alternative methods in terms of PRRs. 
Factuality decoding is hypothesized to increase factual correctness by amplifying knowledge that is localised within particular layers of the language model.
This amplification, however, can distort the probability distribution of the base LLMs, potentially degrading downstream performance.
The results in Table~\ref{tab:rq1-msp-scores} and Table~\ref{tab:rq1-mte-scores} reveal that factuality decoding strategies provide overconfident MSP score and less entropy, suggesting that the original probability distribution of the language model is indeed being altered by emphasising factual tokens.

\begin{figure*}[t!]
    \centering
    \includegraphics[width=15.5cm]{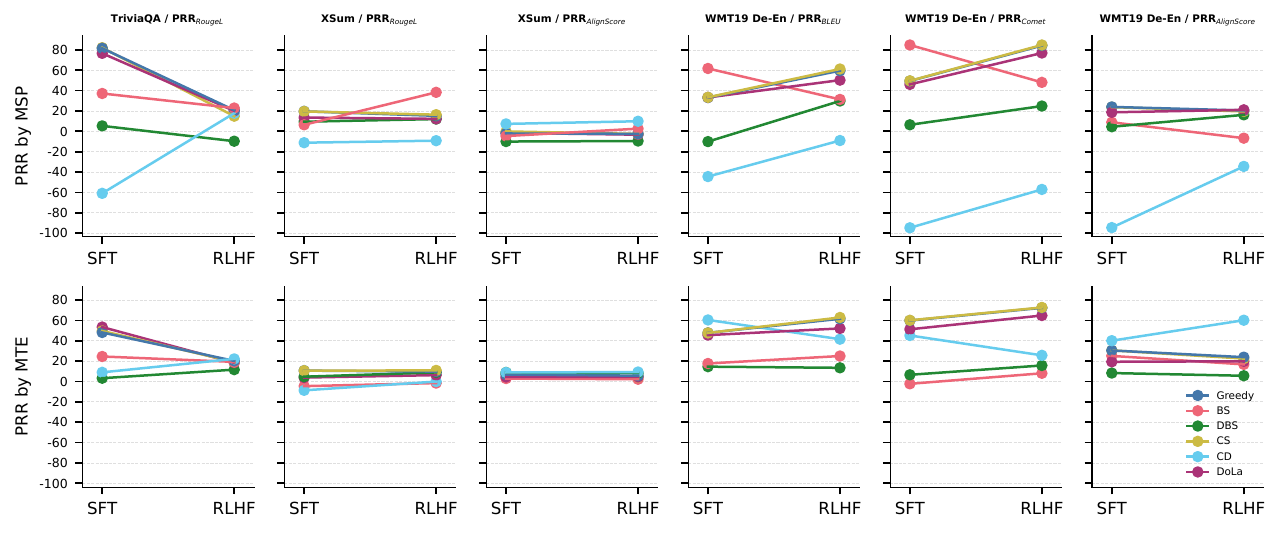}
    \caption{Slopegraphs of PRR when changing the model from Llama3-8B-SFT to Llama3-8B-RLHF.}
    \label{fig:rq2_slopegraphs_sft_rlhf}
\end{figure*}

\begin{figure*}[t!]
    \centering
    \includegraphics[width=15.5cm]{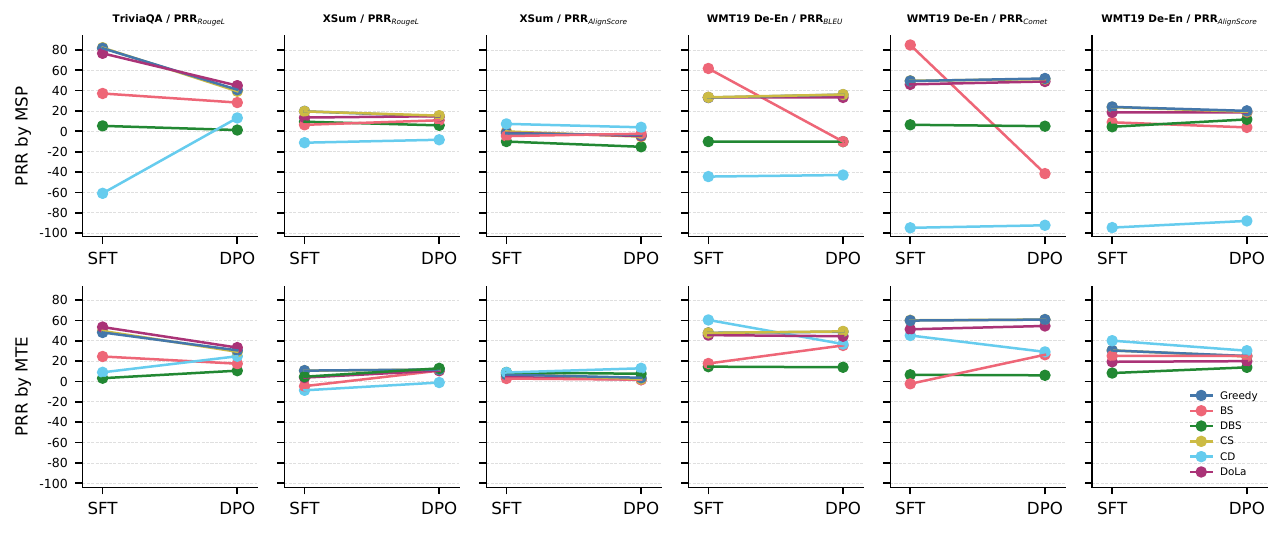}
    \caption{Slopegraphs of PRR when changing the model from Llama3-8B-SFT to Llama3-8B-DPO.}
    \label{fig:rq2_slopegraphs_sft_dpo}
\end{figure*}

\subsection{RQ2: How do training stages such as SFT and the preference-alignment techniques modulate UE performance across decoding strategies?}
\paragraph{Stopping post-train at the SFT stage may affect the conclusion of RQ1.}
Figure~\ref{fig:rq2_slopegraphs_sft_rlhf} depicts the change of PRR values across all task-quality pairs when the training phase of Llama3-8B is switched from SFT to RLHF.
The figure reveals that under SFT, BS achieves superior PRR in a larger number of cases than it does under RLHF.
Previous research~\cite{Kumar2019CalibrationOE} has shown that the confidence calibration effect of BS has positive impacts not only the confidence but also the generation quality.
Furthermore, applying RLHF to an SFT model tends to make its token-level probability distribution more overconfident~\cite{xie-etal-2024-calibrating}.
Consequently, during beam search, low uncertainty score can be assigned to low quality outputs.
This miscalibration will lead to a degradation in PRR.

The absolute impact of the training stage transition on PRR is task-dependent.
On TriviaQA, applying RLHF reduces PRR, whereas on WMT19 De-En it tends to enhance PRR.
Overall, we did not observe the tendency that RLHF induces overconfidence and reduces predictive uncertainty~\cite{kadavath2022languagemodelsmostlyknow,xie-etal-2024-calibrating} when PRR was used as the evaluation metric.

On the other hand, as shown in Figure~\ref{fig:rq2_slopegraphs_sft_dpo}, there are fewer cases in which PRR improved when applied DPO compared to Figure~\ref{fig:rq2_slopegraphs_sft_rlhf}, suggesting overconfident than Llama3-8B-RLHF.
The reason why DPO has lower UE performance compared to RLHF can be due to an interesting ``squeezing effect''~\cite{ren2025learning} in the training of Llama3-8B-DPO.
The ``squeezing effect'' happens a concentration of probability mass on the most likely token by the negative gradient when using DPO-like loss, while Proximal Policy Optimization (PPO)~\cite{schulman2017proximalpolicyoptimizationalgorithms} loss in RLHF avoids the effect~\cite{ren2025learning}.
The phenomenon that Greedy or CS improved PRR in MT on the RLHF stage was not observed on the DPO stage, which can also be due to the increase in the probability of the most likely token and the degeneration of the probability other than the most likely token.

\section{Conclusion}
\label{sec:conclusion}
In this study, we examined how decoding strategies affect predictive uncertainty in LLMs. 
Our experiments show that Contrastive Search strategy tends to provide better uncertainty estimates across various tasks and models on average by mitigating output repetition, a key source of model overconfidence.
On the other hand, we found that the conclusions may change depending on the stage in the post-training phase, such as SFT and the preference-alignment.
We hope that this study will help practitioners improve the reliability of LLMs.

\section*{Limitations}
\paragraph{LLMs}
Our study mainly relies on two aligned models (Llama2-7B-Chat and Llama3-8B-RLHF) and a single SFT model (Llama3-8B-SFT). Experiments on larger models are limited by available resources.
In addition, proprietary models such as GPT4~\cite{openai2024gpt4technicalreport} or Gemini series~\cite{geminiteam2024gemini15unlockingmultimodal,geminiteam2025geminifamilyhighlycapable} are black boxes. Therefore, users cannot freely manipulate the decoding strategy.
All experiments fix the prompt template; we do not explore how prompt engineering might change the conclusions.

\paragraph{Decoding Strategies}
From the viewpoint of practice, we did not consider stochastic decoding strategies, which cannot guarantee deterministic outputs, as discussed in Section~\ref{sec:decoding_strategies}.
In addition, our small-scale experiments do not suggest that stochastic decoding strategies are effective for PRR (see Appendix~\ref{sec:appendix_stochastic_decoding_strategies}).
However, we may find the better stochastic strategy in terms of PRR by more extensive experiments.
Moreover, some recent decoding strategies such as $\phi$-Decoding~\cite{xu-etal-2025-ph} are omitted for limited resource reasons.

\paragraph{UE Methods}
Our analysis focuses on two classical, token-probability based uncertainty estimators - MSP and MTE.
More advanced techniques such as Semantic Entropy~\cite{kuhn2023semantic}, Shifting Attention to Relevance~\cite{duan-etal-2024-shifting}, and distance-based methods~\cite{yoo-etal-2022-detection,hashimoto-etal-2025-efficient} are not benchmarked systematically.
As a result, we cannot claim that the decoding strategy ranking we report would persist when paired with stronger uncertainty estimators.

\paragraph{Tasks}
Our benchmark suite covers four English-only generation tasks with public test sets.
Tasks such as multi-modal understanding~\cite{Yue_2024_CVPR}, combining Retrieval Augmented Generation (RAG) setting~\cite{ozaki-etal-2025-understanding}, and non-English setting~\cite{raihan-etal-2025-mhumaneval} are out of scope.
By improving comprehensiveness, we are likely to gain a deeper understanding of the strengths and weaknesses of each decoding strategy.

\section*{Ethical Considerations}
\paragraph{AI Assistant Tools} We used ChatGPT\footnote{\url{https://openai.com/index/chatgpt/}} and GitHub Copilot\footnote{\url{https://github.com/features/copilot}} to accelerate our research.

\paragraph{Datasets \& Models} This study relies exclusively on publicly available datasets (TriviaQA, XSum, WMT19 De→En, and HumanEval) and openly released LLMs.
All datasets and LLMs used in this study are, at a minimum, licensed for research purposes.
In addition, the datasets we used do not consist of harmful domains (see Appendix~\ref{sec:appendix_dataset_details}).

\paragraph{Uncertainty Estimation} Even a high PRR score can miss low-quality generations; therefore, critical decisions must always include qualified human oversight.

\section*{Acknowledgements}
The authors also acknowledge the Nara Institute of Science and Technology's HPC resources made available for conducting the research reported in this paper.

\bibliography{custom}

\appendix

\section{Details of Decoding Strategies}
\label{sec:appendix_details_decoding_strategies}
\paragraph{Greedy Search (Greedy)} is the simplest decoding strategy, where at each time step $t$ the token with the highest conditional probability is selected. Formally, given an input $\bm{x}$ and the previously generated sequence $\bm{y}_{<t} = \{y_1, y_2, \ldots, y_{t-1}\}$, the next token $y_t$ is chosen as:
\begin{equation}
    y_t = \underset{y \in \mathcal{V}} {\operatorname{argmax}}P(y \mid \bm{y}_{<t}, \bm{x})
\end{equation}
where $\mathcal{V}$ is the vocabulary. 
While Greedy is computationally efficient, its myopic nature may lead to suboptimal overall sequences since only the locally optimal choice is considered at each step.

\paragraph{Beam Search (BS)}~\cite{freitag-al-onaizan-2017-beam} addresses the limitations of Greedy by keeping track of the top-$k$ highest-scoring partial sequences (beams) at each time step.
At step $t$, each beam $\bm{y}_{<t}^{(i)}$ is extended with every possible next token $y\in\mathcal{V}$, producing candidates scored by the cumulative log‐probability:

\begin{equation}
    \mathrm{score}(\bm{y}_{1:t}^{(i)}) 
    = \sum_{\tau=1}^t \log P\bigl(y_\tau \mid \bm{y}_{<\tau}^{(i)}, \bm{x}\bigr).
\end{equation}

Only the top-$k$ candidates are retained as beams for the next time step, trading off exploration and efficiency.
BS often yields higher‐quality sequences than Greedy but can still suffer from low diversity and search errors when $k$ is small.
In this study, we tuned beam size among 3, 5, and 7.

\paragraph{Diverse Beam Search (DBS)}~\cite{Vijayakumar_Cogswell_Selvaraju_Sun_Lee_Crandall_Batra_2018} augments classical beam search with an explicit diversity prior.
All $k$ hypotheses are partitioned into $G$ groups of equal size $k/G$.
At every decoding step, the algorithm first ranks candidates inside each group and then retains the top-$k/G$ sequences per group, rather than the global top-$k$.
The score assigned to a partial sequence $(\bm{y}_{<t},y)$ belonging to group $g$ is

\begin{equation}
\begin{split}
\text{score}\bigl(\bm{y}_{<t},y\bigr)
    &= \log P\bigl(\bm{y}_{<t},y,\big|,\bm{x}\bigr) \\
    &- \lambda \sum_{g' < g}
    \Delta\bigl((\bm{y}_{<t},y), \mathcal{B}_{t}^{g'}\bigr),
\end{split}
\end{equation}

where $\mathcal{B}_{t}^{g'}$ denotes the beam of group $g'$ at time $t$ and
$\Delta(\cdot,\cdot)$ is a similarity measure (e.g., $n$-gram overlap).
DBS encourages beams to explore different regions of the search space, improving output variety.
In this study, we tuned beam size $k$ and group size $G$ among (3, 3), (6, 3), (9, 3), (6, 6), and (12, 6).

\paragraph{Contrastive Search (CS)}~\cite{su2022a,su2023contrastive} assumes that the language model (LM) embeds tokens in an approximately isotropic space.
Given the context $(\bm{x},\bm{y}_{<t})$, it selects the next token by jointly maximizing likelihood and dissimilarity to the preceding hidden states:
\begin{equation}
\begin{split}
y_t &= \underset{y\in\mathcal{V}^k} {\operatorname{argmax}}
    \Bigl[(1-\alpha) P(y\mid \bm{x},\bm{y}_{<t}) \\
    &- \alpha \max_{1 \leq j \leq t - 1}
    s\bigl(h_y,h_{x_{j}}\bigr)\Bigr],
\end{split}
\end{equation}
where $\mathcal{V}^k$ is the top-$k$ candidate set, $h$ are hidden states, and $s$ is usually the cosine similarity. The presence of the second term causes the language model to avoid tokens that are too similar to previous ones, reducing degeneration.
We select the best $\alpha$ among 0.2, 0.4, and 0.6.

\paragraph{Contrastive Decoding (CD)}~\cite{li-etal-2023-contrastive} similarly incorporates a contrastive penalty but directly modifies token‐level logits by using an amateur language model.
For each candidate token $y$:

\begin{align}
    \text{score}\bigl(\bm{y}_{<t},y\bigr) = (1 - \beta) \bm{z}^{e}_{y} - \beta  \bm{z}^{a}_{y},
\end{align}

where $\bm{z}^{e}_{y}$ and $\bm{z}^{a}_{y}$ are logits in the expert language model and the amateur language model, respectively. In addition, CD introduces the following vocabulary constraints to penalize scores by taking into account the grammatical ability and commonsense of the amateur language model:

\begin{equation}
\begin{split}
& \mathcal{V}^{head} \bigl(\bm{x},\bm{y}_{<t}\bigr)
= \{ y \in \mathcal{V}: P^{e} (y \mid \bm{y}_{<t}, \bm{x})\\ & > \alpha \max P^{a} (y \mid \bm{y}_{<t}, \bm{x})\},
\end{split}
\end{equation}

where $P^{e}$ and $P^{a}$ are softmax probability in the expert language model and the amateur language model, respectively.
We set $\alpha = 0.1$, and search $\beta \in \{0.1, 0.3, 0.5, 0.7, 0.9\}$.

\paragraph{Frustratingly Simple Decoding (FSD)}~\cite{yang-etal-2024-frustratingly} contrasts an LM ($P^{\text{base}}$) with an on-the-fly anti-LM ($P^{\text{anti}}$) estimated from the current prefix to penalize the repetition.
Two instantiations exist: an $n$-gram model (FSD) and a vectorized model (FSD-vec).

The selection rule is

\begin{equation}
\begin{split}
    P^{\text{FSD}}(y \mid \bm{y}_{<t}, \bm{x}) & = (1 - \alpha) P^{\text{base}}(y \mid \bm{y}_{<t}, \bm{x}) \\
    &- \alpha P^{\text{anti}} (y \mid \bm{y}_{<t}, \bm{x})
\end{split}
\end{equation}

evaluated over $\mathcal{V}^n$, the top-$n$ tokens under $P^{\text{base}}$.
We tuned $n \in \{3, 5\}$ and $\alpha \in \{0.3, 0.5, 0.7\}$.

\paragraph{Decoding by Contrastive Layers (DoLa)}~\cite{chuang2024dola} is a decoding strategy designed to enhance the factuality of language models.
This method derives a more factual next-token distribution by contrasting the standard next-token prediction obtained from the model's final layer with a prediction from an earlier, or "premature," layer.
Specifically, DoLa utilizes the difference between the logits of the final layer and those of a premature layer to adjust the distribution, thereby encouraging the selection of higher-confidence words.

The selection of the premature layer is dynamic and employs the Jensen-Shannon Divergence (JSD) as a metric to measure the distance between next-token probability distributions.
From a set of candidate premature layers, the one exhibiting the largest JSD with the final layer's probability distribution is chosen.
This approach aims to identify a layer that contains significantly different information compared to the final layer, thereby emphasizing their contrast.
For Llama2-7B and Llama3-8B series, we selected the premature layers from [0, 16) and [16, 32).

\paragraph{Self-Logits Evolution Decoding (SLED)}~\cite{zhang2024sled} improves the factuality of LLM outputs by evolving the model’s logits during decoding to dynamically adjust the token selection process.
SLED achieves this by first contrasting the logits from the model's final layer with those from selected earlier, "premature," layers to unearth potential factual inconsistencies or underexpressed knowledge.
It then employs an approximate gradient-based approach, where this identified latent knowledge guides a "self-evolution" or refinement process of the output probability distribution.
This iterative adjustment aims to steer the generation towards more factually accurate tokens, effectively improving truthfulness while maintaining fluency and incurring negligible latency. 
Consequently, SLED helps LLMs produce more reliable and factually sound text by better aligning their outputs with their inherent knowledge.

In SLED, the main hyperparameters are the top $n$ tokens compared to the logit and the evolution rate $\alpha$ in the logit evolution.
We search $n \in \{5, 10\}$ and $\alpha \in \{0.1, 1.0, 5.0\}$.

\section{Details of Datasets}
\label{sec:appendix_dataset_details}
Dataset statistics are in Table~\ref{tab:dataset_stats}.

\paragraph{TriviaQA}~\cite{joshi-etal-2017-triviaqa} is a large-scale reading-comprehension dataset including question–answer pairs authored independently of evidence documents. 
Each question is paired with supporting context drawn from both Wikipedia and diverse web sources, enabling evaluation of open-domain and extractive QA systems.

\paragraph{XSum}~\cite{narayan-etal-2018-xsum} is a large-scale, single-document abstractive summarization dataset consisting of BBC news articles paired with professionally written, single-sentence summaries.

\paragraph{WMT19}~\cite{wmt19translate} refers to the training and evaluation data released for the 2019 Workshop on Machine Translation shared task, which is designed to benchmark neural machine translation systems.
It includes distinct development and test sets to measure translation for news.

\paragraph{HumanEval}~\cite{chen2021evaluatinglargelanguagemodels} is a collection of 164 programming problems, each paired with a reference implementation and a suite of unit tests.

\begin{table}[t]
\scalebox{0.80}{
\begin{tabular}{lcc}
    \toprule
    Task & Dataset & $N$ \\
    \midrule
    QA & TriviaQA~\cite{joshi-etal-2017-triviaqa} & 17,210 \\
    TS & XSum~\cite{narayan-etal-2018-xsum} & 11,334 \\
    MT & WMT19 (De-En)~\cite{wmt19translate} & 2,998 \\
    CG & HumanEval~\cite{chen2021evaluatinglargelanguagemodels} & 164 \\
    \bottomrule
\end{tabular}
}
\caption{Dataset statistics.}
\label{tab:dataset_stats}
\end{table}

\section{Instruction Templates}
\label{sec:appendix_instruction_template}
The instruction templates for each task are listed from Figure~\ref{fig:prompt_question_answering} to Figure~\ref{fig:prompt_code_generation}.

\begin{figure}[t!]
\begin{minipage}{\columnwidth}
\small
    \begin{mdframed}
    \# Question: \{question\} \\

    \# Answer:
    \end{mdframed}
\end{minipage}
\caption{The prompt for QA.}
\label{fig:prompt_question_answering}
\end{figure}

\begin{figure}[t!]
\begin{minipage}{\columnwidth}
\small
    \begin{mdframed}
    Article: \{text\} \\ \\
    Summarize the above article in 1 sentence. 
    \end{mdframed}
\end{minipage}
\caption{The prompt for TS.}
\label{fig:prompt_text_summarization}
\end{figure}

\begin{figure}[t!]
\begin{minipage}{\columnwidth}
\small
    \begin{mdframed}
    Translate the following sentence from German to English. \\
    \{text\}

    \end{mdframed}
\end{minipage}
\caption{The prompt for MT.}
\label{fig:prompt_machine_translation}
\end{figure}

\begin{figure}[t!]
\begin{minipage}{\columnwidth}
\small
    \begin{mdframed}
    Please complete the remaining Python function code based on the following docstring content.\\
    \{text\}
    \end{mdframed}
\end{minipage}
\caption{The prompt for CG.}
\label{fig:prompt_code_generation}
\end{figure}

\section{Quality Scores}
\label{sec:appendix_quality_scores}
Results for each quality scores are listed in Table~\ref{tab:rq1-quality-scores}.

\begin{table}[t!]
\scalebox{0.45}{

\begin{tabular}{c|l|c|cc|ccc|c}
\hline

\multicolumn{1}{c|}{\textbf{Model}} &
\multicolumn{1}{c|}{\textbf{Method}} &
\multicolumn{1}{c|}{TriviaQA} &
\multicolumn{2}{c|}{XSum} &
\multicolumn{3}{c|}{WMT19} &
\multicolumn{1}{c}{HumanEval} \\ \cline{3-9}
& & \multicolumn{1}{c|}{RougeL} &
\multicolumn{1}{c}{RougeL} & \multicolumn{1}{c|}{AlignScore} &
\multicolumn{1}{c}{BLEU} & \multicolumn{1}{c}{Comet} & \multicolumn{1}{c|}{AlignScore} &
\multicolumn{1}{c}{Pass@1} \\ \hline
\multirow{9}{*}{\rotatebox[origin=c]{90}{Llama2-7B-Chat}}
& Greedy      & 11.36 & 15.08 & 17.43 & 17.44 & 66.62 & 77.68 & 34.76\\
& BS      & 12.16 & 17.82 & 18.88 & 15.31 & 62.17 & 78.38 & 29.88\\
& DBS     & 10.91 & 17.81 & 18.57 & 15.50 & 62.56 & 78.40 & 31.10\\
& CS      & 10.45 & 17.79 & 17.40 & 18.88 & 68.26 & 77.69 & 37.80\\
& CD      &  5.03 & 14.95 & 17.65 & 12.04 & 57.12 & 62.33 & 15.85 \\
& FSD     &  4.98 & 16.97 & 18.01 & 10.49 & 54.72 & 78.10 & 34.76\\
& FSD-vec &  3.37 & 17.06 & 18.40 & 11.29 & 54.68 & 78.27 & 17.07\\
& DoLa    & 11.60 & 17.86 & 17.89 & 17.28 & 66.00 & 77.80 & 36.59\\
& SLED    & 10.91 &  --   &  --   & 14.22 & 63.85 & 78.65 & 46.95\\ \hline
\multirow{6}{*}{\rotatebox[origin=c]{90}{Llama3-8B-RLHF}}
& Greedy & 5.75 & 31.19 & 76.71 & 83.07 & 16.86 & 16.47 & 39.63\\
& BS     & 5.70 & 15.02 & 52.13 & 84.35 & 13.92 & 21.19 & 18.90\\
& DBS    & 5.61 & 15.28 & 52.68 & 84.47 & 14.03 & 21.10 & 25.00\\
& CS     & 5.88 & 30.31 & 75.73 & 83.17 & 16.83 & 16.74 & 42.68\\
& CD     & 5.09 & 6.90  & 54.54 & 57.97 & 17.24 & 18.09 & 28.05\\
& DoLa   & 5.86 & 35.03 & 80.93 & 83.64 & 17.33 & 16.71 & 33.54\\ \hline
\multirow{6}{*}{\rotatebox[origin=c]{90}{Llama3-8B-SFT}}
& Greedy & 53.00 & 20.60 & 10.37 & 40.16 & 86.04 & 83.05 & 48.17 \\
& BS & 17.14 & 20.07 & 12.73 & 18.99 & 56.97 & 85.19 & 7.93 \\
& DBS & 18.11 & 20.11 & 12.53 & 18.90 & 56.94 & 84.98 & 20.12 \\
& CS & 52.50 & 20.46 & 10.55 & 40.11 & 86.01 & 82.78 & 47.56 \\
& CD & 33.16 & 20.99 & 12.38 & 7.14 & 56.40 & 59.47 & 23.78 \\
& DoLa & 55.69 & 21.13 & 10.60 & 41.28 & 86.16 & 83.65 & 49.39 \\ \hline
\end{tabular}
}
\caption{Quality scores for every task and generation metric pair in Llama2-7B-Chat, Llama3-8B-RLHF and Llama3-8B-SFT.}
\label{tab:rq1-quality-scores}
\end{table}

\section{Additional Results on Qwen2.5 Series}
\label{sec:appendix_results_qwen2.5}

Table~\ref{tab:prr-results-qwen2.5-7b} and Table~\ref{tab:prr-results-qwen2.5-14b} present the UE performance achieved with Qwen2.5-7B-Instruct~\cite{qwen2025qwen25technicalreport}\footnote{\url{https://huggingface.co/Qwen/Qwen2.5-7B-Instruct}} and Qwen2.5-14B-Instruct,\footnote{\url{https://huggingface.co/Qwen/Qwen2.5-14B-Instruct}} respectively. For CD, we used Qwen2.5-0.5B-Instruct\footnote{\url{https://huggingface.co/Qwen/Qwen2.5-0.5B-Instruct}} as the amateur model.

\begin{table*}[t!]
\centering
\scalebox{0.47}{

\begin{tabular}{l|c|cc|ccc|c|c||c|cc|ccc|c|c}
\hline
\multicolumn{1}{c|}{\textbf{Method}} &
\multicolumn{8}{c||}{\textbf{MSP}} &
\multicolumn{8}{c}{\textbf{MTE}} \\ \cline{2-17}

\multicolumn{1}{c|}{} &
\multicolumn{1}{c|}{TriviaQA} &
\multicolumn{2}{c|}{XSum} &
\multicolumn{3}{c|}{WMT19} &
\multicolumn{1}{c|}{HumanEval} &
\multicolumn{1}{c||}{} &
\multicolumn{1}{c|}{TriviaQA} &
\multicolumn{2}{c|}{XSum} &
\multicolumn{3}{c|}{WMT19} &
\multicolumn{1}{c|}{HumanEval} &
\multicolumn{1}{c}{} \\ \cline{2-17}

\multicolumn{1}{c|}{} &
\multicolumn{1}{c|}{RougeL} &
\multicolumn{1}{c}{RougeL} & \multicolumn{1}{c|}{AlignScore} &
\multicolumn{1}{c}{BLEU} & \multicolumn{1}{c}{Comet} & \multicolumn{1}{c|}{AlignScore} &
\multicolumn{1}{c|}{Pass@1} & \multicolumn{1}{c||}{Mean PRR} &
\multicolumn{1}{c|}{RougeL} &
\multicolumn{1}{c}{RougeL} & \multicolumn{1}{c|}{AlignScore} &
\multicolumn{1}{c}{BLEU} & \multicolumn{1}{c}{Comet} & \multicolumn{1}{c|}{AlignScore} &
\multicolumn{1}{c|}{Pass@1} & \multicolumn{1}{c}{Mean PRR} \\ \hline
Greedy & \secbest{67.56} & \secbest{10.77} & -0.32 & \secbest{35.33} & \secbest{56.03} & \best{28.55} & \secworst{-9.63} & \secbest{26.90} & \secbest{60.90} & 8.00 & 5.11 & \secbest{43.56} & \secbest{59.06} & \secbest{30.34} & -9.40 & \secbest{28.22} \\
BS & \secworst{62.90} & 10.29 & \secworst{-2.54} & 32.31 & 52.00 & 25.41 & \best{0.36} & 25.82 & 56.84 & \worst{5.99} & \worst{3.26} & 40.43 & 53.98 & 28.06 & \best{0.84} & 27.06 \\
DBS & 63.80 & \secworst{4.42} & \worst{-10.46} & \worst{5.24} & \secworst{29.51} & \secworst{14.28} & -9.27 & \secworst{13.93} & \worst{39.29} & \secworst{6.87} & \secbest{5.34} & \worst{20.08} & \secworst{31.59} & \secworst{10.32} & \secbest{0.64} & \secworst{16.30} \\
CS & \best{67.59} & \best{11.08} & 0.03 & \best{35.36} & \best{56.58} & \secbest{28.38} & -8.15 & \best{27.27} & \best{60.93} & \secbest{8.22} & \best{5.38} & \best{43.60} & \best{59.56} & \best{31.08} & -5.86 & \best{28.99} \\
CD & \worst{-21.87} & \worst{-23.64} & \best{4.11} & \secworst{17.27} & \worst{-32.21} & \worst{-19.28} & \worst{-9.90} & \worst{-12.22} & \secworst{48.36} & \best{11.33} & \secworst{3.84} & \secworst{29.67} & \worst{0.27} & \worst{8.81} & \secworst{-9.84} & \worst{13.21} \\
DoLa & 64.58 & 9.46 & \secbest{3.58} & 34.63 & 54.37 & 24.08 & \secbest{-8.01} & 26.10 & 58.34 & 7.09 & 5.09 & 38.10 & 52.95 & 23.65 & \worst{-10.56} & 24.95 \\ \hline
\end{tabular}
}
\caption{PRRs for every task and generation metric pair in Qwen2.5-7B-Instruct.}
\label{tab:prr-results-qwen2.5-7b}
\end{table*}

\begin{table*}[t!]
\centering
\scalebox{0.47}{

\begin{tabular}{l|c|ccc|c|c||c|ccc|c|c}
\hline
\multicolumn{1}{c|}{\textbf{Method}} &
\multicolumn{6}{c||}{\textbf{MSP}} &
\multicolumn{6}{c}{\textbf{MTE}} \\ \cline{2-13}
\multicolumn{1}{c|}{} &
\multicolumn{1}{c|}{TriviaQA} &
\multicolumn{3}{c|}{WMT19} &
\multicolumn{1}{c|}{HumanEval} &
\multicolumn{1}{c||}{} &
\multicolumn{1}{c|}{TriviaQA} &
\multicolumn{3}{c|}{WMT19} &
\multicolumn{1}{c|}{HumanEval} &
\multicolumn{1}{c}{} \\ \cline{2-13}
\multicolumn{1}{c|}{} & 

\multicolumn{1}{c|}{RougeL} &
\multicolumn{1}{c}{BLEU} & \multicolumn{1}{c}{Comet} & \multicolumn{1}{c|}{AlignScore} &
\multicolumn{1}{c|}{Pass@1} & \multicolumn{1}{c||}{Mean PRR} &
\multicolumn{1}{c|}{RougeL} &
\multicolumn{1}{c}{BLEU} & \multicolumn{1}{c}{Comet} & \multicolumn{1}{c|}{AlignScore} &
\multicolumn{1}{c|}{Pass@1} & \multicolumn{1}{c}{Mean PRR} \\ \hline

Greedy & \secbest{71.44} & \secbest{41.37} & \secbest{67.48} & \best{27.83} & \secworst{2.25} & \secbest{42.07} & \secbest{66.81} & \secbest{47.28} & \secbest{65.93} & \best{31.46} & 6.31 & \best{43.56} \\
BS & 65.55 & 37.89 & 62.51 & 23.18 & \secbest{9.18} & 39.66 & 62.65 & 43.19 & 58.79 & 27.19 & \secbest{13.16} & 41.00 \\
DBS & 68.04 & \worst{13.62} & \secworst{48.34} & \secworst{10.64} & \worst{-7.73} & \secworst{26.58} & \secworst{42.15} & \worst{19.45} & \secworst{31.45} & \worst{12.08} & \worst{-7.94} & \secworst{19.44} \\
CS & \best{71.63} & \best{41.60} & \best{67.85} & \secbest{27.35} & 5.18 & \best{42.72} & \best{66.90} & \best{47.53} & \best{66.37} & \secbest{30.97} & \secworst{4.69} & \secbest{43.29} \\
CD & \worst{-9.90} & \secworst{17.63} & \worst{21.30} & \worst{-19.60} & 8.98 & \worst{3.68} & \worst{18.89} & \secworst{22.28} & \worst{8.87} & \secworst{13.14} & 11.95 & \worst{15.03} \\
DoLa & \secworst{65.00} & 39.96 & 65.36 & 25.84 & \best{13.31} & 41.89 & 58.94 & 41.14 & 62.21 & 26.52 & \best{13.17} & 40.40 \\ \hline
\end{tabular}
}
\caption{PRRs for every task and generation metric pair in Qwen2.5-14B-Instruct.}
\label{tab:prr-results-qwen2.5-14b}
\end{table*}

\section{Additional Results on Llama3-13B-Chat}
\label{sec:appendix_results_larger_model}
Table~\ref{tab:prr-results-13b} presents the UE performance achieved with Llama2-13B-Chat.\footnote{\url{https://huggingface.co/meta-llama/Llama-2-13b-chat-hf}}

\begin{table*}[t!]
\centering
\scalebox{0.47}{

\begin{tabular}{c|c|ccc|c|c||c|ccc|c|c}
\hline
\multicolumn{1}{c|}{\textbf{Method}} &
\multicolumn{6}{c||}{\textbf{MSP}} &
\multicolumn{6}{c}{\textbf{MTE}} \\ \cline{2-13}
\multicolumn{1}{c|}{} &
\multicolumn{1}{c|}{TriviaQA} &
\multicolumn{3}{c|}{WMT19} &
\multicolumn{1}{c|}{HumanEval} &
\multicolumn{1}{c||}{} &
\multicolumn{1}{c|}{TriviaQA} &
\multicolumn{3}{c|}{WMT19} &
\multicolumn{1}{c|}{HumanEval} &
\multicolumn{1}{c}{} \\ \cline{2-13}
\multicolumn{1}{c|}{} & 

\multicolumn{1}{c|}{RougeL} &
\multicolumn{1}{c}{BLEU} & \multicolumn{1}{c}{Comet} & \multicolumn{1}{c|}{AlignScore} &
\multicolumn{1}{c|}{Pass@1} & \multicolumn{1}{c||}{Mean PRR} &
\multicolumn{1}{c|}{RougeL} &
\multicolumn{1}{c}{BLEU} & \multicolumn{1}{c}{Comet} & \multicolumn{1}{c|}{AlignScore} &
\multicolumn{1}{c|}{Pass@1} & \multicolumn{1}{c}{Mean PRR} \\ \hline

Greedy & \secbest{57.08} & \secworst{32.92} & \worst{49.87} & \secworst{29.15} & -17.94 & 30.22 & \best{39.44} & 45.92 & 46.89 & 31.25 & -16.58 & 29.38 \\
BS & 56.27 & 43.8 & 62.42 & 30.28 & -16.8 & 35.19 & 32.22 & 33.37 & 26.25 & \secworst{21.47} & -16.44 & 19.37 \\
DBS & 53.92 & \worst{-44.82} & 60.03 & \worst{18.71} & \secbest{4.57} & \worst{18.48} & \worst{8.52} & \worst{-34.45} & \worst{-66.69} & \worst{-17.64} & \secbest{4.29} & \worst{-21.19} \\
CS & \best{57.85} & 41.55 & 60.64 & 29.33 & \best{15.89} & \best{41.05} & \secbest{38.0} & 43.59 & 45.38 & 31.09 & \best{4.65} & 32.54 \\
CD & \worst{6.43} & 42.49 & \secworst{56.25} & 59.3 & \worst{-20.76} & \secworst{28.74} & \secworst{11.99} & 50.68 & 62.55 & 65.63 & \worst{-25.72} & 33.03 \\
FSD & 20.56 & \best{56.52} & \secbest{62.52} & \secbest{70.69} & -13.01 & \secbest{39.46} & 20.78 & \best{57.71} & \secbest{63.04} & \secbest{70.52} & -11.55 & \best{40.1} \\
FSD-vec & \secworst{19.86} & \secbest{56.15} & \best{64.03} & \best{70.82} & -14.51 & 39.27 & 20.23 & \secbest{57.36} & \best{64.55} & \best{70.78} & -13.03 & \secbest{39.98} \\
DoLa & 55.64 & 44.0 & 61.33 & 31.38 & \secworst{-20.25} & 34.42 & 30.92 & \secworst{32.92} & \secworst{23.44} & 23.95 & \secworst{-20.77} & \secworst{18.09} \\ \hline
\end{tabular}
}
\caption{PRRs for every task and generation metric pair in Llama2-13B-Chat.}
\label{tab:prr-results-13b}
\end{table*}

\begin{table*}[t!]
\centering
\scalebox{0.47}{

\begin{tabular}{c|c|ccc|c|c||c|ccc|c|c}
\hline
\multicolumn{1}{c|}{\textbf{Method}} &
\multicolumn{6}{c||}{\textbf{MSP}} &
\multicolumn{6}{c}{\textbf{MTE}} \\ \cline{2-13}
\multicolumn{1}{c|}{} &
\multicolumn{1}{c|}{TriviaQA} &
\multicolumn{3}{c|}{WMT19} &
\multicolumn{1}{c|}{HumanEval} &
\multicolumn{1}{c||}{} &
\multicolumn{1}{c|}{TriviaQA} &
\multicolumn{3}{c|}{WMT19} &
\multicolumn{1}{c|}{HumanEval} &
\multicolumn{1}{c}{} \\ \cline{2-13}
\multicolumn{1}{c|}{} & 
RougeL &
BLEU & Comet & AlignScore &
Pass@1 & Mean PRR &
RougeL &
BLEU & Comet & AlignScore &
Pass@1 & Mean PRR \\ \hline
Greedy & 62.97 & 38.74 & 46.48 & 19.02 & -11.03 & 31.24 & 49.13 & 31.24 & 25.03 & 21.69 & -13.49 & 22.72 \\ \hline
Temperature & 63.73 & 38.45 & 46.19 & 19.35 & -13.92 & 30.76 & 51.22 & 30.15 & 23.69 & 21.34 & -14.00 & 22.48 \\ 
Top-$p$ & 61.82 & 38.65 & 46.79 & 19.00 & -11.34 & 30.98 & 51.16 & 30.92 & 26.01 & 21.27 & -13.32 & 23.21 \\ \hline
\end{tabular}
}
\caption{PRRs for every task and generation metric pair in Llama2-7B-Chat with Temperature Sampling and Top-$p$ sampling.}
\label{tab:prr-results-stochastic-decoding}
\end{table*}

\section{Experiments on Stochastic Decoding Strategies}
\label{sec:appendix_stochastic_decoding_strategies}
To succinctly evaluate the uncertainty impact of the stochastic decoding strategies omitted from our comprehensive experiments in Section~\ref{sec:decoding_strategies}, we experimented on Temperature Sampling ($T \in \{0.8,1.0,1.2\}$) and Top-$p$ Sampling~\cite{Holtzman2020The} ($p=0.9$).
The results in Table~\ref{tab:prr-results-stochastic-decoding} show that UE performance remains nearly identical to Greedy, suggesting that introducing stochasticity confers little reliability benefit.

\begin{table}[t!]
\centering
\scalebox{0.65}{

\begin{tabular}{c|c|ccc}
\hline
\multicolumn{1}{c|}{Method} &
\multicolumn{1}{c|}{TriviaQA} &
\multicolumn{3}{c}{WMT19} \\ \cline{2-5}
\multicolumn{1}{c|}{} & 
RougeL &
BLEU & Comet & AlignScore \\ \hline
Greedy-MSP & 62.97 & 38.74 & 46.48 & 19.02 \\
CS-MSP & 63.73 & 36.94 & 41.99 & 19.58 \\ 
FSD-MSP & 33.84 & 31.82 & 14.04 & 8.15 \\
DoLa-MSP & 61.15 & 38.78 & 49.15 & 16.74 \\ 
\hline
Greedy-TokenSAR & 51.77 & 29.26 & 24.11 & 21.85 \\
CS-TokenSAR & 51.45 & 30.83 & 26.93 & 23.45 \\
FSD-TokenSAR & -2.14 & 58.36 & 59.46 & 63.68 \\
DoLa-TokenSAR & 50.19 & 28.16 & 23.40 & 15.32 \\ \hline
\end{tabular}
}
\caption{PRRs for every task and generation metric pair in Llama2-7B-Chat with TokenSAR~\cite{duan-etal-2024-shifting}.}
\label{tab:prr-results-tokensar}
\end{table}

\section{Additional Results on Advanced UE Method}
\label{sec:appendix_advanced_ue_method}
We combine TokenSAR, a variant of Shifting Attention to Relevance~\cite{duan-etal-2024-shifting}, with each decoding strategy and show the results in Table~\ref{tab:prr-results-tokensar}. $\text{PRR}_{AlignScore}$ scores from Greedy and CS, and FSD-TokenSAR in MT setting outperform MSP, while the rest degrade.
Existing benchmarking~\cite{10.1162/tacl_a_00737} that comprehensively investigated UE performance has shown that simple MSP is superior, and these results are consistent with those of the previous study.

\begin{table}[t!]
\centering
\scalebox{0.45}{

\begin{tabular}{c|c|c|cc|ccc|c}
\hline
\multicolumn{1}{c|}{Model} & \multicolumn{1}{c|}{Method} &
\multicolumn{1}{c|}{TriviaQA} &
\multicolumn{2}{c|}{XSum} &
\multicolumn{3}{c|}{WMT19} &
\multicolumn{1}{c}{HumanEval} \\ \cline{3-9}
\multicolumn{1}{c|}{} & \multicolumn{1}{c|}{} & 
RougeL &
RougeL & AlignScore &
BLEU & Comet & AlignScore &
Pass@1 \\ \hline
\multirow{9}{*}{\rotatebox[origin=c]{90}{Llama2-7B-Chat}}
& Greedy & - & - & - & - & - & - & - \\
& BS & 3        & 3        & 7        & 3        & 3        & 5        & 3        \\
& DBS & 9\_3   & 9\_3   & 6\_3   & 6\_3   & 3\_3     & 3\_3     & 9\_3     \\
& CS & 0.2      & 0.2      & 0.2      & 0.6      & 0.6      & 0.6      & 0.6      \\
& CD & 0.7      & 0.5      & 0.5      & 0.3      & 0.1      & 0.1      & 0.5      \\
& FSD & 5\_0.7 & 3\_0.3    & 3\_0.5    & 3\_0.3    & 3\_0.3    & 5\_0.5    & 5\_0.3    \\
& FSD-vec & 5\_0.5   & 3\_0.3    & 5\_0.5    & 3\_0.5    & 3\_0.5    & 3\_0.5    & 3\_0.5    \\
& DoLa & [16, 32) & [0, 16)     & [16, 32) & [0, 16)  & [0, 16)  & [0, 16)  & [16, 32) \\
& SLED & 5.0\_5   & 1.0\_5    & 1.0\_5    & 0.1\_5    & 5.0\_10  & 0.1\_5    & 5.0\_5   \\
\hline
\multirow{6}{*}{\rotatebox[origin=c]{90}{Llama3-8B-RLHF}}
& Greedy & - & - & - & - & - & - & - \\
& BS & 7 & 3 & 5 & 3 & 3 & 3 & 3 \\
& DBS & 9\_3 & 3\_3 & 9\_3 & 12\_6 & 3\_3 & 6\_3 & 3\_3 \\
& CS & 0.6 & 0.2 & 0.4 & 0.2 & 0.2 & 0.2 & 0.6 \\
& CD & 0.3 & 0.5 & 0.5 & 0.5 & 0.5 & 0.5 & 0.1 \\
& DoLa & [16, 32) & [16, 32) & [16, 32) & [0, 16) & [16, 32) & [16, 32) & [0, 16) \\ \hline
\multirow{7}{*}{\rotatebox[origin=c]{90}{Zephyr-7B-$\beta$}}
& Greedy & - & - & - & - & - & - & - \\
& BS & 7 & 7 & 5 & 3 & 3 & 3 & 5 \\
& DBS & 9\_3 & 9\_3 & 9\_3 & 9\_3 & 9\_3 & 3\_3 & 3\_3 \\
& CS & 0.2 & 0.2 & 0.4 & 0.4 & 0.6 & 0.4 & 0.4 \\
& FSD & 3\_0.3 & 3\_0.3 & 3\_0.3 & 5\_0.3 & 5\_0.5 & 5\_0.5 & 5\_0.5 \\
& FSD-vec & 3\_0.3 & 3\_0.3 & 3\_0.3 & 5\_0.3 & 5\_0.3 & 5\_0.5 & 3\_0.5 \\
& DoLa & [0, 16)  & [0, 16) & [0, 16) & [0, 16) & [0, 16) & [16, 32) & [0, 16) \\ \hline
\end{tabular}
}
\caption{Optimal hyperparameters in Table~\ref{tab:rq1-prr-results}.}
\label{tab:rq1-prr-results-hyperparameters}
\end{table}

\begin{table}[t!]
\centering
\scalebox{0.45}{
\begin{tabular}{c|c|cc|ccc|c}
\hline
\multicolumn{1}{c|}{Method} & 
\multicolumn{1}{c|}{TriviaQA} &
\multicolumn{2}{c|}{XSum} &
\multicolumn{3}{c|}{WMT19} &
\multicolumn{1}{c}{HumanEval} \\ \cline{2-8}
\multicolumn{1}{c|}{} & 
RougeL &
RougeL & AlignScore &
BLEU & Comet & AlignScore &
Pass@1 \\ \hline
Greedy & - & - & - & - & - & - & - \\
BS & 7 & 3 & 5 & 3 & 3 & 3 & 3 \\
DBS & 9\_3 & 3\_3 & 9\_3 & 12\_6 & 3\_3 & 6\_3 & 3\_3 \\
CS & 0.6 & 0.2 & 0.4 & 0.2 & 0.2 & 0.2 & 0.6 \\
CD & 0.3 & 0.5 & 0.5 & 0.5 & 0.5 & 0.5 & 0.1 \\
DoLa & [16, 32) & [16, 32) & [16, 32) & [0, 16) & [16, 32) & [16, 32) & [0, 16) \\ \hline
\end{tabular}
}
\caption{Optimal hyperparameters in Llama3-8B-SFT.}
\label{tab:sft-prr-results-hyperparameters}
\end{table}

\begin{table}[t!]
\centering
\scalebox{0.45}{
\begin{tabular}{c|c|cc|ccc|c}
\hline
\multicolumn{1}{c|}{Method} & 
\multicolumn{1}{c|}{TriviaQA} &
\multicolumn{2}{c|}{XSum} &
\multicolumn{3}{c|}{WMT19} &
\multicolumn{1}{c}{HumanEval} \\ \cline{2-8}
\multicolumn{1}{c|}{} & 
RougeL &
RougeL & AlignScore &
BLEU & Comet & AlignScore &
Pass@1 \\ \hline
Greedy & - & - & - & - & - & - & - \\
BS & 7 & 3 & 5 & 5 & 5 & 5 & 3 \\
DBS & 9\_3 & 9\_3 & 9\_3 & 6\_3 & 9\_3 & 6\_3 & 9\_3 \\
CS & 0.6 & 0.4 & 0.2 & 0.4 & 0.2 & 0.2 & 0.6 \\
CD & 0.1 & 0.5 & 0.7 & 0.7 & 0.3 & 0.5 & 0.5 \\
FSD & 5\_0.7 & 5\_0.5 & 5\_0.5 & 3\_0.5 & 5\_0.5 & 5\_0.3 & 3\_0.7 \\
FSD-vec & 3\_0.7 & 5\_0.5 & 3\_0.5 & 5\_0.3 & 3\_0.7 & 5\_0.3 & 5\_0.5 \\
DoLa & [0, 16) & [0, 16) & [16, 32) & [0, 16) & [0, 16) & [16, 32) & [0, 16) \\ \hline
\end{tabular}
}
\caption{Optimal hyperparameters in Table~\ref{tab:prr-results-qwen2.5-7b}.}
\label{tab:prr-results-hyperparameters-qwen2.5-7b}
\end{table}

\begin{table}[t!]
\centering
\scalebox{0.45}{
\begin{tabular}{c|c|ccc|c}
\hline
\multicolumn{1}{c|}{Method} & 
\multicolumn{1}{c|}{TriviaQA} &
\multicolumn{3}{c|}{WMT19} &
\multicolumn{1}{c}{HumanEval} \\ \cline{2-6}
\multicolumn{1}{c|}{} & 
RougeL &
BLEU & Comet & AlignScore &
Pass@1 \\ \hline
Greedy & - & - & - & - & - \\
BS & 7 & 5 & 5 & 7 & 5 \\
DBS & 9\_3 & 9\_3 & 9\_3 & 9\_3 & 6\_6 \\
CS & 0.2 & 0.2 & 0.2 & 0.2 & 0.4 \\
CD & 0.1 & 0.1 & 0.3 & 0.3 & 0.5 \\
DoLa & [16, 32) & [0, 16) & [0, 16) & [0, 16) & [16, 32) \\ \hline
\end{tabular}
}
\caption{Optimal hyperparameters in Table~\ref{tab:prr-results-qwen2.5-14b}.}
\label{tab:prr-results-hyperparameters-qwen2.5-14b}
\end{table}

\begin{table}[t!]
\centering
\scalebox{0.45}{
\begin{tabular}{c|c|ccc|c}
\hline
\multicolumn{1}{c|}{Method} & 
\multicolumn{1}{c|}{TriviaQA} &
\multicolumn{3}{c|}{WMT19} &
\multicolumn{1}{c}{HumanEval} \\ \cline{2-6}
\multicolumn{1}{c|}{} & 
RougeL &
BLEU & Comet & AlignScore &
Pass@1 \\ \hline
Greedy & - & - & - & - & - \\
BS & 3 & 3 & 3 & 3 & 5 \\
DBS & 6\_3 & 6\_3 & 6\_3 & 6\_6 & 6\_6 \\
CS & 0.2 & 0.6 & 0.6 & 0.4 & 0.6 \\
CD & 0.1 & 0.1 & 0.1 & 0.1 & 0.1 \\
DoLa & [0, 16) & [0, 16) & [0, 16) & [0, 16) & [16, 32) \\ \hline
\end{tabular}
}
\caption{Optimal hyperparameters in Table~\ref{tab:prr-results-13b}.}
\label{tab:prr-results-hyperparameters-13B}
\end{table}

\section{Details of Implementation}
\label{sec:appendix_details_implementation}
We used a single NVIDIA A100 40GB for all experiments.
Decoding strategies have been implemented with reference to Hugging Face Transformers~\cite{wolf-etal-2020-transformers} and official implementations.\footnote{\url{https://github.com/XiangLi1999/ContrastiveDecoding}}\footnote{\url{https://github.com/LHRYANG/FSD}}\footnote{\url{https://github.com/JayZhang42/SLED/}}
Quality metrics, uncertainty metrics, and uncertainty estimation methods have been implemented with reference to LM-polygraph~\cite{fadeeva-etal-2023-lm}.

\section{Settings of Hyperparameters}
\label{sec:appendix_optimized_hyperparameters}
The optimal hyperparameters for each decoding strategy across different datasets and models are listed from Table~\ref{tab:rq1-prr-results-hyperparameters} to Table~\ref{tab:prr-results-hyperparameters-13B}.




\end{document}